# CXR-AD: Component X-ray Image Dataset for Industrial Anomaly Detection

Haoyu Bai[†], Jie Wang[†], Gaomin Li, Xuan Li, Xiaohu Zhang,  and Xia Yang[‡]

**Abstract**—Internal defect detection constitutes a critical process in ensuring component quality, for which anomaly detection serves as an effective solution. However, existing anomaly detection datasets predominantly focus on surface defects in visible-light images, lacking publicly available X-ray datasets targeting internal defects in components. To address this gap, we construct the first publicly accessible component X-ray anomaly detection (CXR-AD) dataset, comprising real-world X-ray images. The dataset covers five industrial component categories, including 653 normal samples and 561 defect samples with precise pixel-level mask annotations. We systematically analyze the dataset characteristics and identify three major technical challenges: (1) strong coupling between complex internal structures and defect regions, (2) inherent low contrast and high noise interference in X-ray imaging, and (3) significant variations in defect scales and morphologies. To evaluate dataset complexity, we benchmark three state-of-the-art anomaly detection frameworks (feature-based, reconstruction-based, and zero-shot learning methods). Experimental results demonstrate a 29.78% average performance degradation on CXR-AD compared to MVTec AD, highlighting the limitations of current algorithms in handling internal defect detection tasks. To the best of our knowledge, CXR-AD represents the first publicly available X-ray dataset for component anomaly detection, providing a real-world industrial benchmark to advance algorithm development and enhance precision in internal defect inspection technologies.

## I. INTRODUCTION

**A**nomaly Detection is a technique used to identify data points, events, or patterns that deviate significantly from the norm. Anomalies, also referred to as outliers, are observations that differ markedly from other data. Although the definition of an anomaly can vary across different application scenarios, it generally refers to instances that occur with extremely low frequency and deviate substantially from the regular pattern or the majority of normal data.In the semiconductor manufacturing process, defects such as internal cavities, foreign materials, or missing components can arise within electronic components. X-ray inspection systems are commonly used to capture images of semiconductor components, and applying anomaly detection techniques to these images enables the identification and analysis of such defects. This process assists inspectors in determining whether there are potential quality issues within the semiconductor devices.

In the field of industrial anomaly detection, high-quality datasets serve as a fundamental cornerstone for advancing algorithmic research and improving model performance [1]. Due to the rarity and difficulty of obtaining anomalous samples in industrial settings, constructing representative datasets is especially crucial for effectively training and evaluating anomaly detection models.

As shown in Fig. 1, defect images in typical industrial scenarios primarily focus on surface defects of objects, which are generally large in scale, exhibit clear boundaries, and have significant contrast with the background. However, for defects located within internal structures, there is currently a lack of high-quality datasets to adequately support the development of anomaly detection algorithms. This limitation hinders the effectiveness and generalization capability of such algorithms in practical applications.

The main contribution of this paper is the introduction of the first dataset specifically designed for X-ray images of semiconductor components. Collected from real-world industrial production environments, this dataset aims to provide defect samples from a new perspective for anomaly detection algorithms, and to serve as a new benchmark for evaluating such algorithms. Ultimately, it seeks to promote the application and development of anomaly detection techniques in the context of industrial defect inspection.

The proposed dataset poses several key challenges for anomaly detection tasks:

1) Complex backgrounds and subtle anomalies: The CXR-AD dataset contains a large number of complex background patterns and subtle anomalous regions, which demand strong local feature modeling capabilities from detection models;

2) Multi-scale anomaly distribution: The dataset features anomalies with a wide range of scales, encompassing both tiny local defects and larger anomalous areas;

3) High detection difficulty: X-ray images inherently suffer from low overall contrast, low grayscale values, and

This paragraph of the first footnote will contain the date on which you submitted your paper for review, which is populated by IEEE. It is IEEE style to display support information, including sponsor and financial support acknowledgment, here and not in an acknowledgment section at the end of the article. For example, "This work was supported in part by the U.S. Department of Commerce under Grant 123456." *(Corresponding author: Xia Yang).* Haoyu Bai and Jie Wang contributed equally to this manuscript.

The authors are with College of Aeronautics and Astronautics, Sun Yat-sen University, Guangdong 510275, China (e-mail: baihy9@mail2.sysu.edu.cn; wangj688@mail2.sysu.edu.cn; ligm23@mail2.sysu.edu.cn; zhangxiaohu@mail.sysu.edu.cn; yangxia7@mail.sysu.edu.cn;).

The fourth author is with the China Aerospace Aerospace Component Engineering Center, Beijing 100080, China (e-mail: lixuan2066@163.com).

significant mixed noise. These factors result in a low signal-to-noise ratio and make it difficult to distinguish defects from the background.

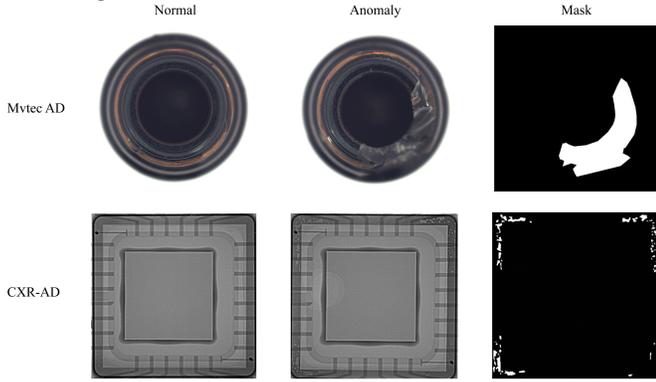

**Fig. 1.** Examples of typical industrial inspection scenarios and component X-ray inspection scenarios.

The remainder of this paper is organized as follows: In the next section, we provide an overview of commonly used anomaly detection datasets and discuss mainstream visual anomaly detection methods. Section III focuses on the data acquisition process of our proposed dataset and presents a quantitative analysis to reveal the characteristics of X-ray images of semiconductor components. Finally, in Section IV, we conduct a comprehensive evaluation of several state-of-the-art anomaly detection algorithms on our dataset, comparing their performance across multiple metrics.

## II. RELATED WORK

### A. Existing anomaly detection datasets

In anomaly detection research, the choice of dataset is crucial, as it directly affects both model training outcomes and evaluation standards. Different datasets typically encompass a variety of application scenarios and defect types, making the selection of an appropriate dataset essential for accurately assessing the performance and applicability of detection algorithms.In recent years, with the growing demand for industrial visual inspection, several anomaly detection datasets targeting different industrial contexts have been introduced. These datasets cover a broad spectrum of defect types, ranging from surface flaws to structural anomalies, providing researchers with rich experimental platforms.In this section, we present several representative datasets commonly used in anomaly detection, highlighting their key characteristics, application domains, and their influence on model training and evaluation.

As one of the benchmark datasets for industrial anomaly detection, MVTec AD [2] is characterized by its high-quality, pixel-level anomaly annotations, making it well-suited for surface defect detection and localization tasks. However, the dataset includes relatively simple types of anomalies, which may not fully capture the complexity of defects encountered in real-world industrial environments. Moreover, its image acquisition settings are highly idealized, lacking the complex backgrounds and noise typically present in practical industrial scenarios.

The VisA [3] dataset contains images of 12 categories of industrial products, covering various materials and surface types. Similar to MVTec AD, VisA provides pixel-level annotations, but it features a greater diversity of anomaly types, making it more suitable for studying anomaly detection in complex scenarios. However, some categories in VisA have relatively few anomalous samples, which may result in suboptimal model performance on those specific categories.

MPDD [4] focuses on defect detection in metal part manufacturing, offering over 1,000 images with pixel-level defect annotations. It is specifically designed for defects on metal surfaces, such as scratches and dents, and is well-suited for high-precision inspection of metallic components. This specialization, however, limits its applicability to other types of materials, making it less generalizable across different industrial domains.

CAD-SD is a dataset tailored to the detection of local and co-occurring anomalies in screws and their accessories. Captured using high-precision industrial cameras, it includes 400 normal images and 376 anomalous images. The anomaly types include local defects (e.g., scratches, paint issues) and co-occurring anomalies (e.g., over-coupling, missing parts). Nevertheless, the image acquisition environment is highly idealized, lacking realistic industrial background complexity and interference. Additionally, the training set contains only 400 normal images, and the evaluation set has a limited number of anomalous samples, which may lead to overfitting or insufficient generalization during model training and evaluation.

### B. Anomaly Detection Methods

In the field of industrial defect detection, the mainstream anomaly detection methods usually have the following three categories: reconstruction-based anomaly detection methods, feature-embedding-based methods, and zero or few-shot anomaly detection methods. The core idea of reconstruction-based methods is to train a model to reconstruct a normal image, while feature-embedding-based methods compare the feature embedding of the target image with that of the normal image to generate pixel-level anomaly maps, and zero or few-shot anomaly detection methods usually introduce a multimodal large language model for anomaly detection.

**Reconstruction-based approach:** in this approach, the model builds a reconstruction process by learning the features of a normal sample. When a normal sample is input, the model is able to accurately reconstruct the original data of that sample, whereas when an anomalous sample is input, the reconstruction error of the model increases significantly due to the large difference between that sample and the normal pattern learned during the training process. Reconstruction-based anomaly detection methods are usually trained using techniques such as Autoencoders or Generative Adversarial Networks (GANs) [5]. Autoencoders map the input to the latent space through an encoder, and then a decoder reconstructs the latent representation back to the original image, whereas GANs consist of a generator and a discriminator, through which samples as close to the real as possible are created, and a discriminator is used to discriminate between true and false samples. The advantage of reconstruction-based methods is that they can effectively capture the structure and features of normal samples, and are effective in detecting anomalous samples that are significantly different from normal samples.

Bergmann *et al.* [6] introduced structural similarity (SSIM) [7] to train the generative model, and the structural similarity loss function takes into account both illumination, contrast, and structural information; however, such methods still measure anomalous regions at a low level, which makes it difficult for high-level semantic anomalies to be extracted in a complete manner. Fang *et al.* [8] proposed a method named FastRecon, which combines fast feature reconstruction and disparity detection techniques on the basis of a self-encoder. Efficient reconstruction in feature mapping space is used to learn the distribution of normal samples and locate abnormal regions. A sample less adaptation strategy is introduced to improve the generalization ability in sample insufficient scenarios. Nguyen *et al.* [9] conducted a comparative analysis of current variational auto-encoder (VAE) architectures applied in anomaly detection, and VAE combined with a visual transformer (ViT-VAE) [10] performs excellently in several scenarios, whereas VAE with a Gaussian random field prior (VAE- GRF) [11], on the other hand, may require more complex hyperparameter tuning to achieve optimal performance. In addition, the MiAD [12] dataset is introduced for benchmarking to avoid over-reliance on the widely used MVTec dataset.

In terms of Generative Adversarial Networks (GANs), AnoGAN [13] is the first method that introduces GAN into defect detection by learning the distribution of normal samples through GAN, then mapping the samples with defects to the hidden variables, and then reconstructing the samples from the hidden variables; since GAN only learns the distribution of the normal samples, the reconstructed image will eliminate the defective parts while retaining the characteristics of the original image; Finally, the location of defects is determined by the residual difference between the reconstructed image and the original image [14] . In order to solve the problem of scarcity of defective images and limited performance of traditional detection methods in industrial scenes, Duan *et al.* [15] proposed a data enhancement method based on StyleGAN2 to generate new defective samples based on a small number of defective images and a relatively large number of normal images. Experiments on the MVTec AD dataset show that the method not only generates realistic and diverse defect images, but also effectively improves the performance of downstream defect detection tasks. The latest research in generative adversarial networks is a rough knowledge-based adversarial learning method proposed by Fang *et al.* [16], which effectively suppresses the self-encoder's ability to reconstruct anomalous samples by aligning the reconstructed feature distributions with the normal feature distributions, thus improving the detection accuracy, and further proposes an image-block-based adversarial learning strategy based on image blocks.

The development process of reconstruction-based methods is actually a continuous improvement of the complexity and performance of the reconstruction sub-network. However, the core goal of this component (the reconstruction subnetwork) remains unchanged: it aims to reconstruct the abnormal image into the corresponding normal image, and to recover the abnormal part of the abnormal image as reasonably as possible. The essence of this approach comes from the idea of reconstruction based on normal samples, i.e., to achieve effective detection of abnormal images by minimizing the reconstruction error of the normal image and maximizing the difference in reconstruction error between the normal and abnormal images.

**Feature-embedding-based methods:** detect anomalies by transforming an image into a high dimensional feature space and thus comparing feature distances between images. These methods usually utilize pre-trained networks (e.g., convolutional neural networks) to extract high-level features of an image and use these features for image-level anomaly detection [17] . Since the pre-trained models have learned rich information about image features, feature-embedding-based methods tend to have better performance in image-level anomaly detection, especially in identifying global, structural anomalies, and can better utilize global contextual information.

Defard *et al.* [18] proposed PaDiM, an anomaly detection and localization framework based on image block distribution modeling, which extracts image block features using pre-trained CNNs and models the probability distribution of normal samples through multivariate Gaussian distribution, while combining with the correlation of CNNs at multi-semantic levels to enhance the anomaly localization accuracy. Roth *et al.* [19] proposed an automated defect detection and localization using only normal samples PatchCore, an automated detection method suitable for multiple tasks, achieves efficient defect detection and localization by constructing a feature library of maximally representative normal image blocks and combining embedded features from ImageNet pre-trained models with an anomaly detection model. Existing methods usually utilize pre-trained visual representations of natural image datasets and extract relevant features, with large discrepancies between the pre-trained features and the target data, Hyun *et al.* [20] constructed discriminative features for anomaly detection by training linear modulations of image block features extracted by the pre-trained model, and used comparison table learning to collect and assign features to produce a target-oriented and easily separable representation. And two similarity measures between data representations: pairwise similarity and contextual similarity are utilized as pseudo-labels to address the lack of labeled pairs for contrast learning.

Pre-trained models typically learn rich feature representations by training on large-scale datasets, which enables them to exhibit better performance in a variety of downstream tasks. In contrast, reconstruction models trained from scratch lack such a priori knowledge, they require more training data and longer training time, and struggle to achieve the robustness that pre-trained models can provide, and thus feature-embedding-based approaches tend to outperform reconstruction-based approaches in terms of image-level performance.

**Zero-Shot and few-Shot based approaches:** existing studies leave much to be desired in terms of model generalization ability when facing unknown class tasks, and the sparsity of sample features makes it impossible to robustly handle complex targets. With the development of large models and multimodality, these methods also play an important role in the field of anomaly detection, and the anomaly detection performance can be improved by designing a generalized detection model or with the help of multimodality.

Inspired by ChatGPT [21], Gu *et al.* [22] proposed a new method AnomalyGPT based on Large Visual Language Model (LVLM) by generating simulated anomalous images and their corresponding textual descriptions as training data, combining with an image decoder to provide fine-grained semantics, and designing a cue learner to fine-tune the LVLM through cue embedding. AnomalyGPT can directly determine the presence and location of anomalies without manually adjusting thresholds, while supporting multiple rounds of dialogs and demonstrating excellent sample-scarce context learning capabilities. Matcher [23] generates initial cue points by combining bi-directional feature matching with DINOv2 [24], and employs a robust sampling strategy to extract multiple sets of cues to guide the segmentation of SAMs. The method demonstrates excellent generalization ability in multiple segmentation tasks by incorporating the advantages of different underlying models. In terms of zero-shot anomaly detection, most of the approaches nowadays are usually investigated based on pre-trained visual language models (CLIP) [25]. AdaCLIP proposed by Cao *et al.* [26] optimizes the CLIP model by introducing learnable cues (prompts), including static cues and dynamic cues: static cues are shared across all images and are used to initially adapt CLIPs to support zero-shot anomaly detection; dynamic prompts are generated for each test image, giving the model the ability to adapt dynamically.

With their strong generalization ability, large models and multimodal provide new solution ideas for zero and few shot segmentation problems, and a series of breakthroughs have been achieved. However, current research still faces some challenges: large models usually use large-scale natural domain images in the pre-training stage, and when there is a large difference in image distribution with the downstream task, the effect of direct migration application is often unsatisfactory; multimodal methods are better at capturing global semantic information, and in complex scenarios such as low-contrast and small targets (e.g., industrial defect detection or medical lesion segmentation), the accuracy of the existing methods is still to be Improvement.

## III. Dataset Description

The Chip X-ray anomaly detection dataset comprises five categories, including 559 images for training, and 655 images for testing. The training set contains only normal images, while the test set includes both normal and defective samples. Table I provides an overview of each category in the dataset. The five categories represent typical semiconductor microfocus chip packaging types, such as CDIP, UB, and others. The defects present in the test images are primarily bubble defects that occur during the chip packaging process. These anomalies are real-world defects captured from actual industrial production environments.

TABLE I
NUMBER OF TRAINING SAMPLES, NORMAL TEST SAMPLES AND ABNORMAL TEST SAMPLES FOR EACH CLASS IN THE DATASET

| Class | Train | Test Normal | Test Anomaly |
|---|---|---|---|
| CDIP | 94 | 13 | 164 |
| CFP16 | 54 | 10 | 60 |
| CFP20 | 135 | 31 | 185 |
| UB | 196 | 20 | 62 |
| DC | 80 | 20 | 90 |
| Total | 559 | 94 | 561 |

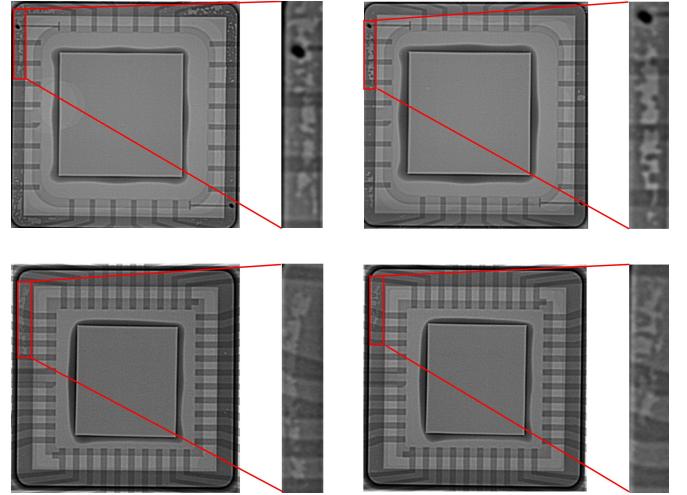

**Fig. 2.** Examples of X-ray image defects and localized enlargements

The X-ray images of components are shown in Fig. 2. All images were captured using the YXLON Cheetah microfocus X-ray inspection system, which employs advanced microfocus X-ray tube and flat-panel detector technologies with sub-micron resolution. This enables the system to clearly detect microscopic defects, such as pores, cracks, and inclusions, both within and on the surfaces of industrial workpieces, including metals and composites. With a dynamic range of up to 16-bit, the system can simultaneously capture fine details in both high-density areas (e.g., metal structures) and low-density regions (e.g., plastics or air gaps) in a single exposure. This ensures that the image contrast and signal-to-noise ratios meet the rigorous standards required for industrial inspection. All image acquisition processes are conducted in a controlled laboratory environment with constant temperature and humidity, effectively mitigating the impact of environmental factors on image quality.

In the experimental data preparation stage, the labeling process is the core of the dataset construction. The LabelMe tool is used to label the polygonal contours of defective areas on a pixel-by-pixel basis, and the vertex coordinates need to be adjusted manually for small and densely distributed defects, and the vertex fine-tuning needs to be repeated 5-8 times for a single sample to match the actual edge of the defects, and a JSON file is generated for each annotation result with the coordinate information; the JSON annotations are then batch converted into binary mask images using a Python script. Subsequently, the JSON labeled data are batch converted to binary mask images by Python script, and the coordinate-pixel mapping algorithm is implemented using OpenCV library to ensure that the defective areas are accurately labeled as the foreground with a pixel value of 1, and the non-defective areas are uniformly set to 0. At the same time, an automated checking module is developed to verify the geometrical consistency between mask boundaries and the original images.

To systematically characterize the inherent challenges of the images in this dataset, we evaluate the X-ray images of semiconductor components using the UB model chip as an example. Specifically, we employ histogram statistics, local contrast measurements, and defect scale analysis to assess the X-ray images.

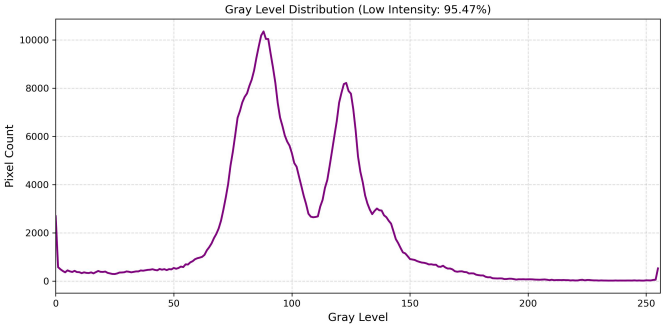

**Fig. 3.** Global grayscale histogram

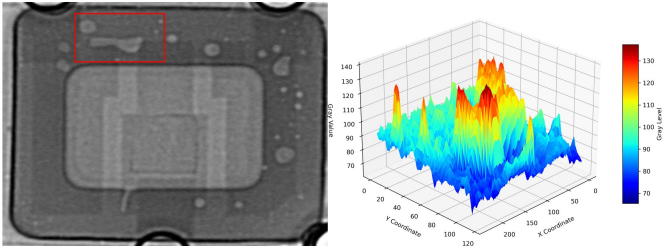

**Fig. 4.** X-ray image of the UB model chip and its three-dimensional gray peak map of the defective region

Gray-level Histogram is a statistical chart that describes the distribution of pixel gray values in an image, reflecting the overall contrast, brightness and dynamic range characteristics of the image. Its mathematical definition is as follows:

$$H(i) = \sum_{(x,y) \in I} \delta(I(x,y) - i) \tag{1}$$

Where $H(i)$ is the number of pixels at gray level $i$ and $I(x,y)$ is the gray value at coordinates $(x,y)$, a Kronecker delta function that takes 1 if $I(x,y)=i$ and 0 otherwise.

As shown in Figs. 3 and 4, by analyzing the characteristics of the grayscale distribution of the defects in the real dataset, the following features can be summarized: 95.47% of the pixel values are concentrated in the range of 50-150, the overall grayscale distribution conforms to the characteristics of the Gaussian distribution, which is decreasing from the center to the surroundings, and there is a transition region at the edges; and there is a large difference in the grayscale of defects spanning across different layered structures inside the defects. This indicates that the gray scale distribution of the X-ray image is skewed towards the lower gray scale regions, and the overall brightness is low. This results in a lack of detail and blurring of the edges of the defects.

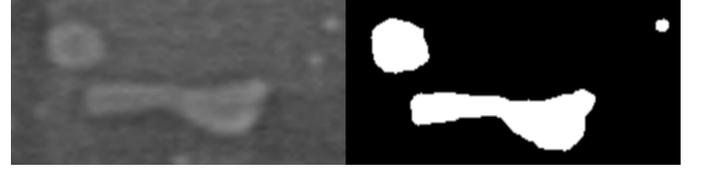

**Fig. 5.** Defective area and mask

Local contrast (LC) is an image processing metric used to measure the difference in intensity between light and dark areas within a localized region of an image. It can be applied to quantify the grayscale difference between a defective region and its surrounding background, and is commonly used in industrial inspection, medical image analysis, and other fields, particularly for low-contrast X-ray images. Fig. 5 shows the defective region and the corresponding mask image. When calculating local contrast (LC), the mask is used to distinguish the defective area from the background. The closer the LC value is to 0, the smaller the grayscale difference between the target area (e.g., defect) and the background.

Local contrast (LC) is calculated as:

$$LC = \frac{\mu_{obj} - \mu_{bg}}{\mu_{bg}} \tag{2}$$

Where $\mu_{obj}$ is the mean gray scale value of the target region (defect) and $\mu_{bg}$ is the mean gray scale value of the background. the closer the $LC$ value is to 0, the smaller the gray scale difference between the defective region and the background.

As shown in Fig. 6, the statistical distribution of local contrast (LC) based on 62 samples of the UB model indicates that the defective regions in semiconductor X-ray images exhibit extremely low contrast characteristics. The LC values fluctuate within the range of 0.01 to 0.22, with an average of 0.1091. This suggests that the grayscale of the defective areas is close to that of the background, making anomaly detection significantly more challenging.

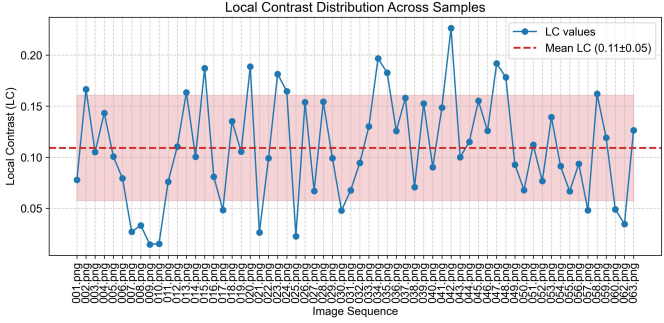

**Fig. 6.** Distribution of local contrast statistics

In terms of defect scale, an adaptive binning method was employed to dynamically analyze the defect scale distribution. The defects were categorized into small, medium, and large groups based on 1%, 10%, and 50% of the maximum defect area. Fig. 7 shows the defect distribution across different scale ranges, revealing that small defects account for 87% of the total, while medium and large defects together make up 12%. Additionally, Fig. 8 presents the area distribution of all defects, with the majority of defects concentrated on the left side, while a few large defects appear on the right. This indicates that the dataset spans a wide range of defect scales, from tiny local defects to larger anomaly regions, which adds complexity to the anomaly detection task.

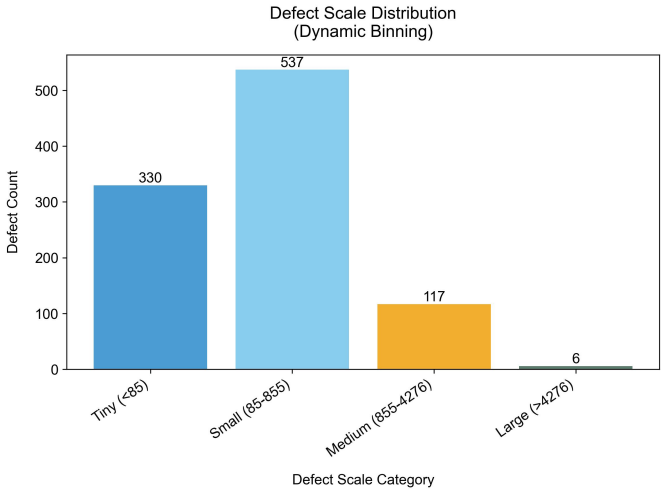

**Fig. 7.** Distribution of local contrast statistics

## IV. BENCHMARK

We conducted a comprehensive evaluation of several advanced unsupervised anomaly detection methods, using them as the initial benchmark for the CXR-AD dataset and as a reference for future methods. The experiments show that although each method is capable of detecting certain types of anomalies, none of them perform exceptionally well across the entire dataset.

The experiments are conducted on a workstation with an Intel Core i9-10920X, 128GB RAM and NVIDIA GeForce RTX 3090Ti with 24GB memory. The software environment is Ubuntu 18.04, Pytorch 1.12.1 and Python 3.8.

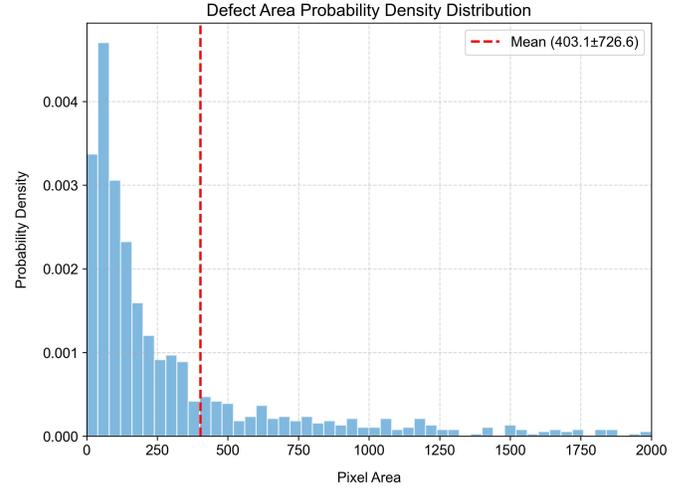

**Fig. 8.** Area distribution of defects

In this paper, image AUROC and pixel AUROC are used as indicators to evaluate the detection effect of different methods.AUROC(Area Under the Receiver Operating Characteristic) is the area under the receiver characteristic curve, i.e., the area under the ROC(Receiver Operating Characteristic Curve) curve. Characteristic Curve) curve. Image AUROC is concerned with whether the entire image contains anomalies (image-level anomaly recognition capability); pixel AUROC is concerned with the model's ability to recognize anomalies at the pixel level.

### A. Feature-embedding-based Methods

In the context of feature-embedding-based Methods, PatchCore was used as a benchmark for testing. The core idea behind PatchCore is to construct a "memory bank of normal features" and compare the features of the test sample with those in the memory bank to detect anomalies. Before inputting the images into the PatchCore model, all images undergo a unified preprocessing step. The images are resized to a fixed size (256×256 pixels) and center-cropped to ensure consistency in the input images. Additionally, no data augmentation is applied to avoid any bias introduced by augmentation. The experimental results show that PatchCore exhibits significant performance variation across different categories, with its metrics being much lower than those obtained when tested on the MVTec AD dataset.

### B. Reconstruction-based Methods

ONENIP [27] is an efficient and compact anomaly detection framework that draws inspiration from predictive coding theory and utilizes normal images as global cues to guide feature reconstruction. The input image resolution is set to 320×320, with EfficientNet-b4 used as the feature extractor. For unsupervised reconstruction or recovery, the number of layers in both the encoder and decoder is set to 4, balancing performance and computational cost. All other settings remain consistent with those in the original paper. Although the model performs moderately well on global metrics, it is significantly limited in detection at the pixel level and suffers from boundary blurring of anomalous regions as well as leakage of small-size defects as shown in the visualization in Fig. 9.

### C. Zero-Shot based method

In the context of Zero-Shot Anomaly Detection, the state-of-the-art multimodal anomaly detection model AdaClip was used as the benchmark for testing. AdaCLIP introduces hybrid learnable prompts and a hierarchical semantic fusion (HSF) module, enabling the detection of anomalies in new classes without the need for training samples of the target class. The model was trained on the VisA and ColonDB datasets and tested on the CXR-AD and MVTec AD datasets. The results show that the CXR-AD dataset presents significant challenges in pixel-level anomaly detection. Due to the small size of the anomalous regions and the complexity of the background, the model's localization ability is significantly reduced, with pixel-level metrics much lower than those on the MVTec AD dataset.

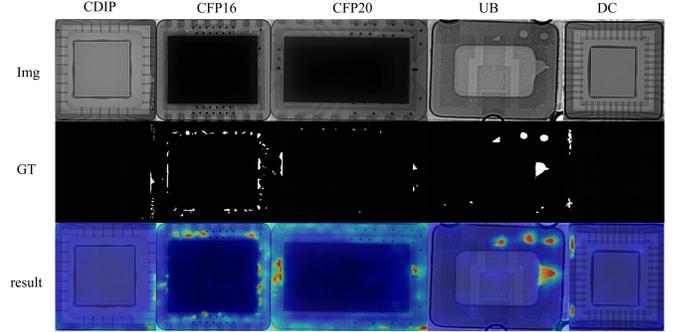

**Fig. 9.** ONENIP visualization results

### V. Conclusion

This study addresses the key challenges in detecting internal defects in semiconductor components by constructing the first CXR-AD dataset for semiconductor anomaly detection based on X-ray imaging. The dataset is systematically analyzed to reveal its inherent characteristics and the challenges it poses to existing algorithms, providing a new benchmark for research on weak textures and multi-scale industrial defect detection. Through histogram statistics, local contrast measurements, and defect scale analysis, the study uncovers the inherent properties of X-ray images, such as low contrast, complex noise distribution, and the coexistence of multi-scale defects, offering theoretical support for the challenges presented by this dataset. The limitations of current mainstream anomaly detection methods on this dataset were also validated, particularly in terms of insufficient local feature modeling ability in complex backgrounds and poor robustness to multi-scale anomalies, providing clear directions for future research. This research not only introduces a new data resource and research perspective for the industrial defect detection field but also reveals the technical bottlenecks of existing methods in handling complex industrial scenarios. It lays the foundation for advancing non-destructive testing technologies based on X-ray imaging.

#### TABLE II
#### Image/Pixel-Level AUROC on CXR-AD Dataset by Different Methods

| Class | Methods | | |
|---|---|---|---|
| | Patchcore | ONENIP | AdaCLIP |
| CDIP | 40.95/89.55 | 98.78/95.62 | 21.93/55.90 |
| CFP16 | 54.08/74.67 | 84.44/89.69 | 50.57/74.48 |
| CFP20 | 32.17/65.50 | 60.73/89.23 | 71.86/78.85 |
| DC | 56.28/49.42 | 83.38/95.72 | 58.31/72.64 |
| UB | 99.43/94.78 | 99.36/98.07 | 73.75/72.14 |
| Avg | 56.58/74.78 | 85.34/93.66 | 55.32/70.80 |
| Avg MvtecAD | 99.00/98.00 | 97.90/97.90 | 89.71/89.90 |

This experiment compares the performance differences of various methods on the CXR-AD and MVTec AD datasets, validating the challenge posed by CXR-AD as a benchmark for industrial X-ray image anomaly detection. The experimental results show that the overall performance of all compared methods on CXR-AD is significantly lower than their performance on MVTec AD. PatchCore's average Image/Pixel AUROC on CXR-AD decreased by more than 40% and 23%, respectively, compared to its performance on MVTec AD. While ONENIP achieved the highest metrics in the experiment, its performance still lags behind MVTec AD by 12.6% and 4.3%. As the most advanced multimodal anomaly detection model, AdaClip underperforms compared to the other two models, suggesting that it may rely heavily on precise textual prompts for specific detection tasks. The introduction of CXR-AD provides a more practical evaluation benchmark for anomaly detection, and the performance gap not only highlights the limitations of existing algorithms in complex defect detection but also offers optimization directions for future research.